\documentclass{article}
\usepackage{graphicx} 
\graphicspath{ {./images/} }

\usepackage{amsmath} 
\usepackage{geometry}
\usepackage{xcolor}
\usepackage{float}

\def \doctype {paper}

\makeatletter
\DeclareRobustCommand{\change}{%
\@bsphack
\normalcolor 
\@esphack
}
\DeclareRobustCommand{\stopchange}{%
\@bsphack
\normalcolor
\@esphack
}
\DeclareRobustCommand{\bc}{%
\@bsphack
\normalcolor 
\@esphack
}
\DeclareRobustCommand{\ec}{%
\@bsphack
\normalcolor
\@esphack
}
\DeclareRobustCommand{\bcc}{%
\@bsphack
\normalcolor 
\@esphack
}
\DeclareRobustCommand{\ecc}{%
\@bsphack
\normalcolor
\@esphack
}

\makeatother

\title{\LARGE \bf

Adaptive Landmark Color for AUV Docking in Visually Dynamic Environments
}

\author{Corey Knutson$^{1}$, Zhipeng Cao$^{2}$, and Junaed Sattar$^{3}$
\thanks{This work was supported by the National Science Foundation Award IIS-2220956. 
The authors are with the Department of Computer Science \& Engineering, University of Minnesota - Twin Cities, Minneapolis, MN, USA. {\tt\small \{$^{1}$knuts983, $^{2}$cao00223, $^{3}$junaed\}@umn.edu}
Accepted for publication at ICRA 2024
}}

\date{}

\begin{document}

\maketitle
\thispagestyle{empty}
\pagestyle{empty}

\begin{abstract}

Autonomous Underwater Vehicles (AUVs) conduct missions underwater without the need for human intervention. 
A docking station (DS) can extend mission times of an AUV by providing a location for the AUV to recharge its batteries and receive updated mission information. Various methods for locating and tracking a DS exist, but most rely on expensive acoustic sensors, or are vision-based, which is significantly affected by water quality. In this \doctype, we present a vision-based method that utilizes adaptive color LED markers and dynamic color filtering to maximize landmark visibility in varying water conditions. Both AUV and DS utilize cameras to determine the water background color in order to calculate the desired marker color. No communication between AUV and DS is needed to determine marker color. Experiments conducted in a pool and lake show our method performs 10 times better than static color thresholding methods as background color varies. DS detection is possible at a range of 5 meters in clear water with minimal false positives.
\end{abstract}

\begin{figure}[h]
    \centering
    \includegraphics[width=8.5cm]{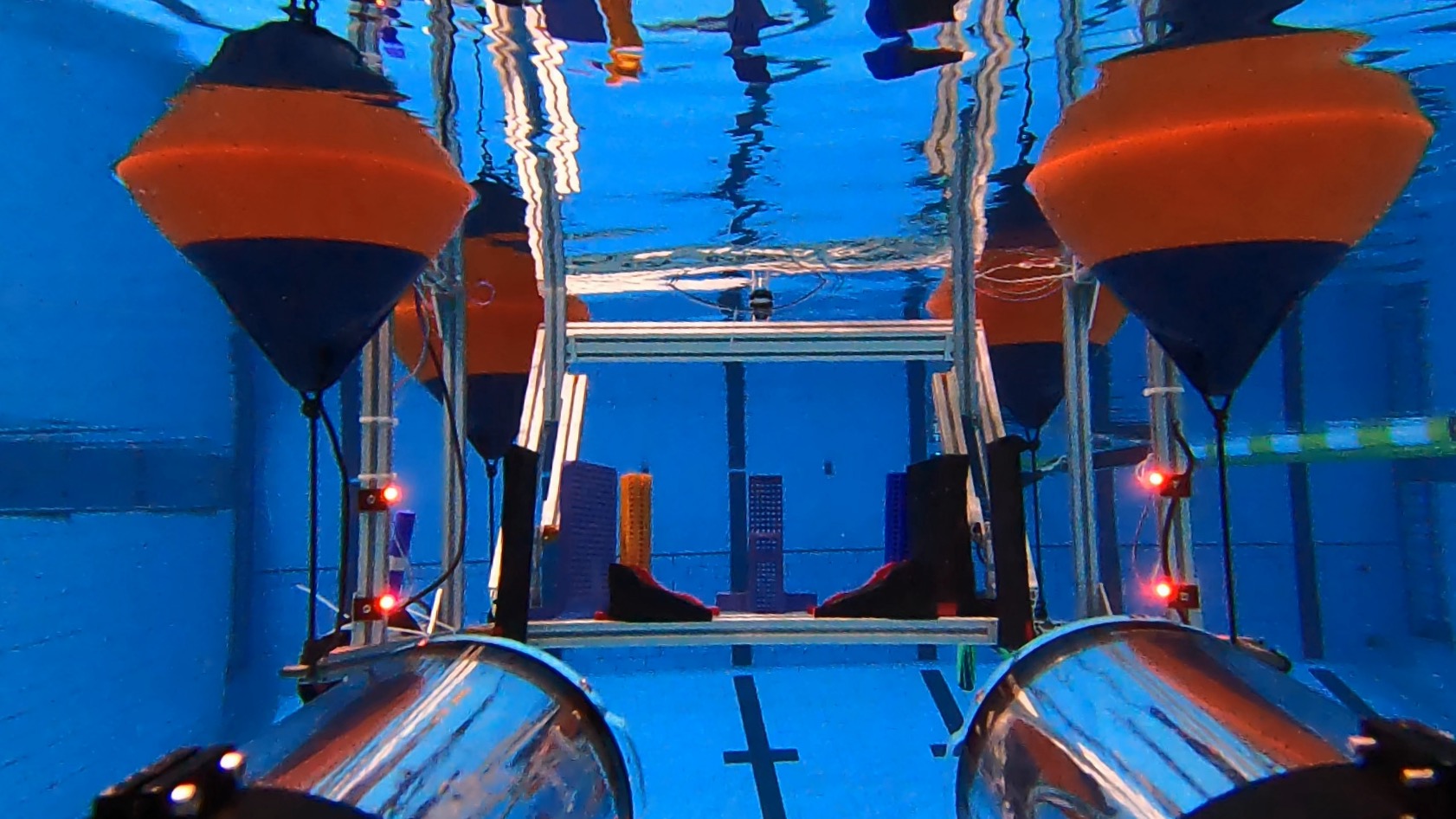}
    \caption{The LoCO AUV (bottom) using adaptive color RGB markers to locate a docking station (top).}
    \label{fig:loco_seacube_approach}
\end{figure}

\section{Introduction}

The capabilities of Autonomous Underwater Vehicles (AUVs) are constantly improving \bc to help humans work safely underwater. \ec AUVs can conduct infrastructure maintenance and inspection \cite{palomeras}, biological sample gathering \cite{fiorelli}, and marine trash detection \cite{fulton-trash} \bc completely autonomously, removing a human from harms way. In more complex missions where human perception and cognition is necessary, a human and robot can communicate with each other underwater to work towards solving a task \cite{fulton-hri}. \ec

A docking station (DS) can significantly extend mission duration by providing charging and communications to an AUV in the field. Two key steps of the AUV docking process are the approach setup, and terminal homing phase \cite{bellingham}. The approach setup is typically realized with acoustic sensors on the AUV and transmitters on the DS, such as USBL, due to their long detection range of over 1km. Using this information, the AUV will make gross movements to position itself in line with the DS entrance, then start its approach. The final 15 meters of the approach is the terminal homing phase, in which the vehicle must accurately detect the relative position of the DS in order to enter the dock at an acceptable trajectory. Acoustic sensors typically lack the level of accuracy required for this maneuver, so many AUVs use cameras to detect active light markers on a DS, which can be used to compute the 3D position, or pose, of the DS. Once docked, the AUV can commence charging and communications, then finally signal the DS to release once the AUV is ready to start its mission.

\bc Submerged \ec docking stations are suitable for large bodies of water where AUVs are operating at or near the depth of the DS. The REMUS, and similar torpedo-style AUVs, use submerged funnel-shaped docking stations anchored to the sea floor \cite{remus}. \bc At depth, \ec nearby water is relatively calm during docking maneuvers. However, in smaller bodies of water, an anchored\bc, \ec floating DS may be favorable. On the surface of the water, a DS can charge an AUV with solar panels and transmit data wirelessly. Additionally, installation and maintenance of the DS and AUV can be performed above the water, which is preferable to submerging and resurfacing a structure from the bottom of the water body. \bcc Fig. \ref{fig:loco_seacube_approach} is an example of a floating DS designed to capture the \bc Low-Cost Open \ec (LoCO) AUV \cite{loco}. \ecc

\begin{figure*}[t]
    \centering
    \includegraphics[width=\textwidth]{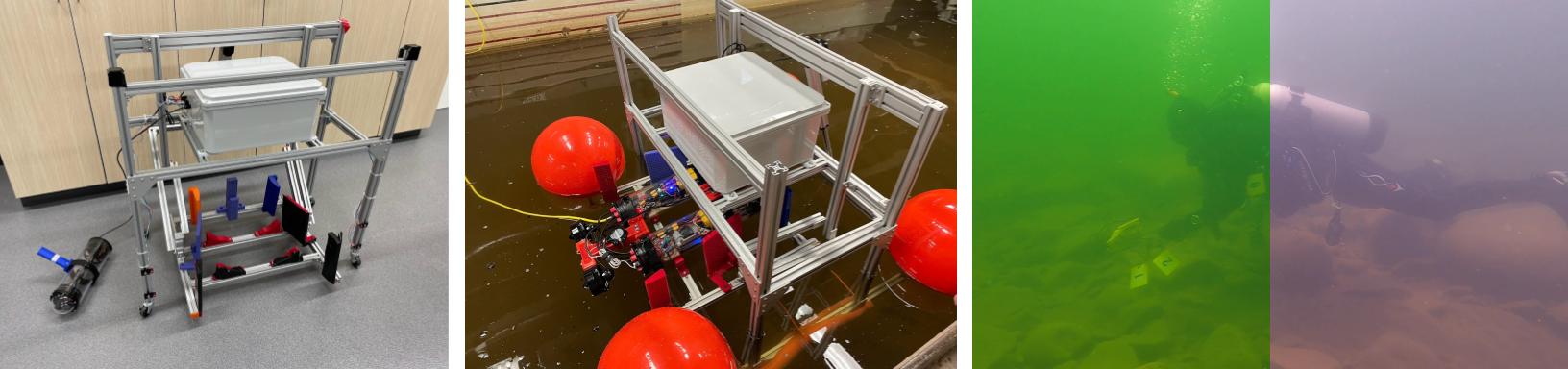}
    \caption{Left - Prototype DS without buoys and detached camera module. Middle - The LoCO AUV successfully docked with the DS in a water flume. Right - A composition of two images taken five seconds apart. A cloud passes overhead, drastically changing water color in Lake Superior.}
    \label{fig:ds_figures_divers}
\end{figure*}

Locating a floating DS during the AUV's approach setup can be simplified as well. The AUV can surface, localize itself using GPS, then receive a position and heading of the DS using wireless communications. A final terminal homing phase is still necessary, but near the surface of the water visual conditions change rapidly depending on ambient lighting and water conditions, which affects the contrast of DS markers, and therefore the accuracy of DS detection systems. \bc Fig. \ref{fig:ds_figures_divers} (right) is a common example of how quickly water color varies. \ec

This \doctype\ proposes a detection system for use with floating docking stations and AUVs that is resistant to background color shifting. As the background color varies, both AUV and DS detect the color change and decide on a marker color that will have high contrast with the background. Two different color mapping functions are proposed and evaluated. The AUV detects the landmarks by color masking with the anticipated landmark color, then performs blob detection and spatial filtering to calculate point correspondences to the known landmark locations. Tests were conducted using the prototype DS pictured in Fig. \ref{fig:ds_figures_divers} and the \bcc LoCO AUV.

\section{Related Works}

\subsection{Color Underwater}

Due to the properties of the medium, underwater imaging can produce pictures that are not true to the subject's actual appearance above water \cite{jaffe}. This is due to a process known as attenuation. There are two main contributors to attenuation: scattering and absorption. The former occurs when light encounters particles suspended in the medium which have similar dimensions as the wavelength of the light, or have a different index of refraction than the surrounding medium. The latter occurs when light intersects with particles and is absorbed, rather than reflected. Attenuation also varies depending on the wavelength of light passing through the medium. In water bodies, lower energy wavelengths such as orange or red (600-700nm) are attenuated much quicker than the rest of the visible spectrum, which causes objects underwater to have a blue-green appearance.

As light is scattered and absorbed, the energy reaching a camera varies depending on the path the light travelled. An underwater imaging model proposed by \cite{jaffe} consists of three main paths the light will follow. The first path is the direct component $E_{d}$, in which the light is not scattered in the water. The second is the forward-scattered component $E_{fs}$, where the light is scattered between the object and the camera. Finally, the back-scattered component $E_{bs}$ is the path in which the light is scattered between its source, background, and the camera. The model in \cite{jaffe} describes the total light energy reaching an imaging sensor as the superposition of these three components and can be described mathematically as

\begin{equation}
E_{T} = E_{d} + E_{fs} + E_{bs}
\label{eqn:superposition_light}
\end{equation}

where $E_{T}$ is the total irradiance.

This model implies that as conditions that affect the $E_{fs}$ and $E_{bs}$ components change, the total light reaching the camera varies in intensity, which changes the appearance of the target in the image. Water turbidity, cloud cover, time of day, algae blooms, depth, and salinity are some of the factors that affect total irridance. Fig. \ref{fig:ds_figures_divers} (right) demonstrates cloud cover drastically varying water color. The diver in these pictures is operating at a depth of 10 meters in Lake Superior.

Underwater Image Enhancement (UIE) methods are being developed to improve image quality and reconstruct the lost appearance of objects in an image. Methods range from deep-learning models, such as FUnIE-GAN \cite{islam}, to physics based approximation of spectrum attenuation \cite{berman} and color restoration techniques \cite{zhang}. While these methods produce visually pleasing results in clear water conditions, ambient lighting is not sufficient to illuminate targets when visibility is reduced by environmental variables, or when light is almost completely absent at night.
\subsection{Detection}

Detection of man-made objects underwater is typically assisted by markers to increase visibility. Marker types can be described as either passive or active. Passive markers are those that only reflect light. A form of passive markers is AprilTags \cite{olson} which can be used to localize a robot with a single marker. The authors of \cite{vivek} use AprilTags to track the pose of a DS and create extra contrast on the tag using headlights mounted on the AUV. Active markers are typically preferred for underwater applications due to light scattering and absorption of natural light. While the light emitted from active markers is still attenuated by the water, the single path of the light from emitter to camera is significantly less affected than the path of natural light.

Existing methods for DS detection include classical image processing techniques, such as image intensity thresholding and blob detection \cite{liu-blob} \cite{trslic} \cite{yahya} \cite{zaman}. An image is first converted from color to greyscale, then a Laplacian of Gaussian or Difference of Gaussian blob detection algorithm can be applied on the image to detect the active landmarks. Sometimes intensity thresholding is applied first to eliminate background noise. While these methods are satisfactory in certain situations, they rely heavily on consistent lighting and background color as the detector is looking for high contrast between the markers and background. If the marker and background color are the same, which is possible in any of these works as the markers are either blue, purple, green, or white, there will be little to no response from the blob detector.
Other marker detection methods include deep learning models, such as \cite{liu-donn}. The network presented in \cite{liu-donn} is robust to surface reflections, bright particle artifacts, and changes in background color, but can still suffer if the DS marker color is too similar to the water color. Additionally, this model must be trained on the marker configuration specific to the DS, which requires substantial amounts of underwater images in diverse lighting conditions.

\section{Methods}

\subsection{Docking station}

Still in its early stages of design and assembly, the prototype DS pictured in Fig. \ref{fig:ds_figures_divers} was used to conduct experiments. This DS floats on the surface of the water and captures an AUV as it enters the DS horizontally. Upon entry into the DS, a proximity sensor will detect the presence of the AUV, which engages a motor to vertically lift the capture carriage, raising the AUV out of the water and settling it into its locked position. Four common-anode RGB LEDs are mounted coplanar to form a rectangle \bc 89cm \ec wide by \bc 12.7cm \ec tall. Upon detection of these landmarks, an AUV can use this prior information to calculate the pose of the DS relative to itself. A \bc RGB camera \ec is mounted in a waterproof tube and attached to a submerged strut on the DS, facing away from the entrance. This camera provides image data for the landmark color calculation which constantly updates the color of the landmarks.

\subsection{Marker Color Choice}

\begin{figure}[h]
    \centering
    \includegraphics[width=7cm]{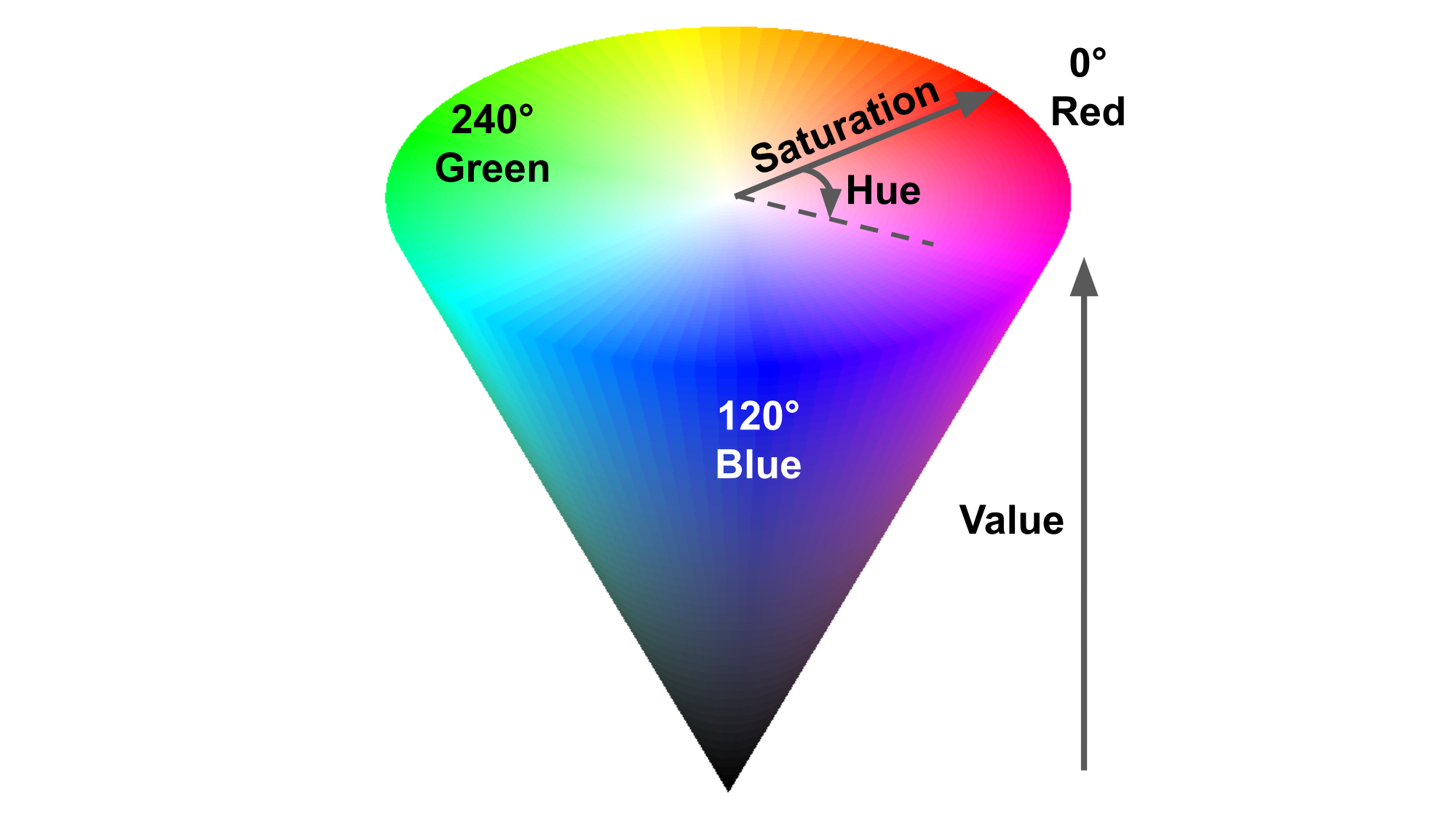}
    \caption{The hexcone model of the HSV color space. \bcc In this model, \textit{chroma} is the radial dimension when value is less than one, not saturation, but we are mainly considering colors whose value is one. \ecc }
    \label{fig:hsvmodel}
\end{figure}

\begin{figure}[t]
    \centering
    \includegraphics[width=7cm]{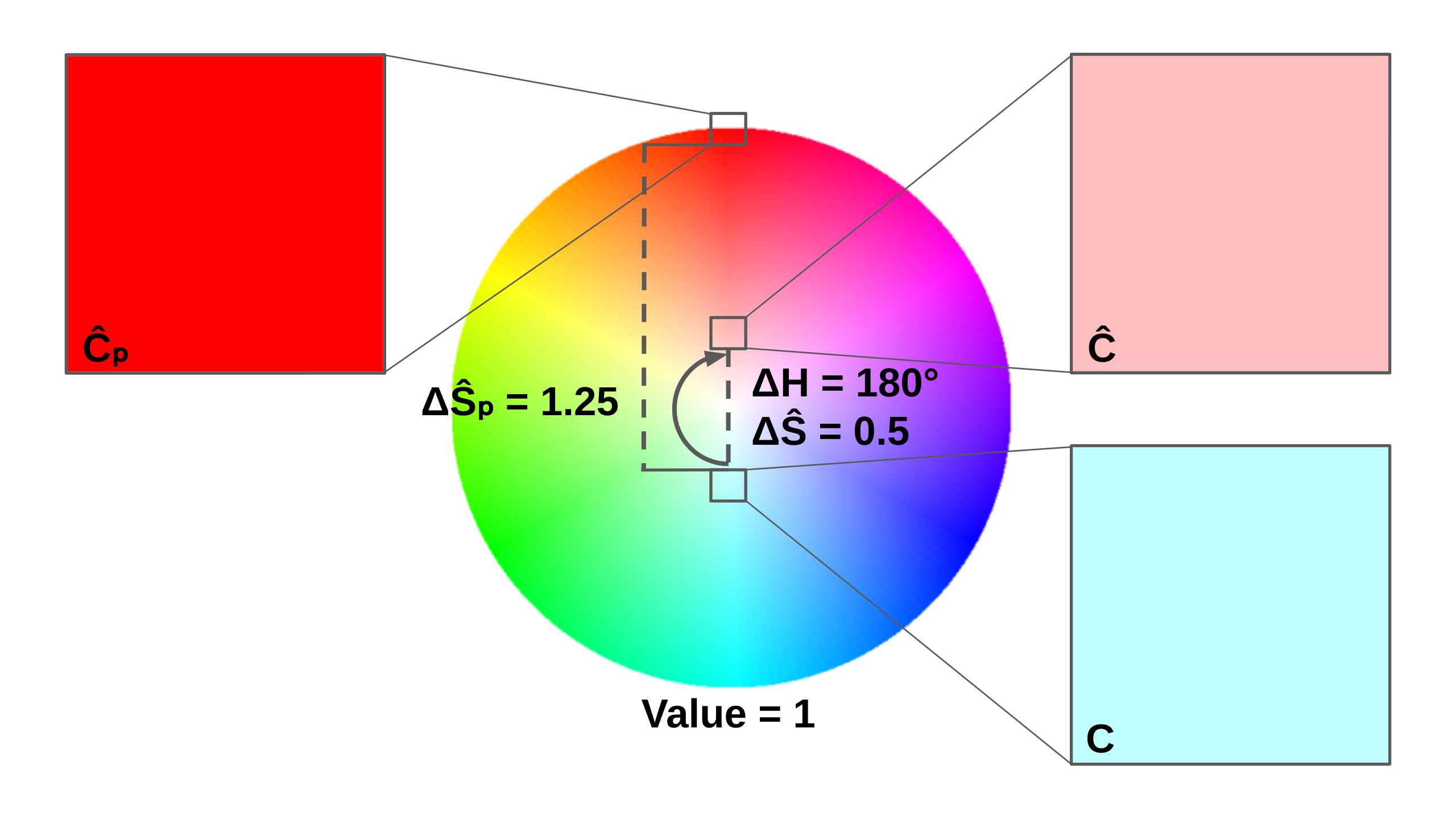}
    \caption{A color $C$, its complement $\widehat{C}$, and pure complement $\widehat{C_{p}}$ in HSV color space.}
    \label{fig:hsv_complement}
\end{figure}

Consider a color $C$ in the Hue Saturation Value (HSV) color space. The color $C$ can be represented by a set of three values: hue $H \in [0, 360]$, saturation $S \in [0, 1]$, and value $V \in [0, 1]$, such that 
\begin{equation}
    C = \{H, S, V\}
\end{equation}
Fig. \ref{fig:hsvmodel} is a visualization of the HSV color space, referred to as the HSV hexcone model \cite{smithhexcone}.

Let hue be an angular dimension in degrees, with red at $0 ^{\circ}$, blue at $120 ^{\circ}$, green at $240 ^{\circ}$, and returning to red at $360 ^{\circ}$. Saturation is a measure of fullness or vibrance of the color and is measured as the length of the radial vector from the center of the cone outward. A lower saturation results in a more \bc ``\ec washed out\bc '' \ec color. Value is a measured from the base of the cone increasing in value vertically, with a lower value resulting in a darker \bc ``\ec shade\bc '' \ec of color. If both value and saturation \bc are maximum\ec, the resulting color can be called a \textit{pure color}. A pure color can be calculated from any HSV color by letting S and V equal one.

For any given color $C$ in the HSV color space, a \textit{complement} $\widehat{C}$ exists such that 
\begin{equation}
    \widehat{C} = \{ |H - 180|, S, V\}
\end{equation}
Since hue is defined in the range $[0, 360]$, the complementary color $\widehat{C}$ must have the highest separation in hue, or contrast of hues, from $C$. However, complementary colors are not always distinguishable from each other. Consider the three colors in Fig. \ref{fig:hsv_complement}. Color $C$ is low in saturation ($S=0.25$), so its complement $\widehat{C}$ is equally low in saturation, with a \textit{saturation difference} $\Delta \widehat{S}=0.5$. In order to maximize separation of both hue and saturation in this color space, a \textit{pure compliment} is defined as 
\begin{equation}
    \widehat{C_{p}} = \{ |H - 180|, 1, 1\}
\end{equation}
It can be seen that $\Delta \widehat{S_{p}} =  \widehat{S} + \widehat{S_{p}} \geq 1$ will hold true for any color $C$ and its pure compliment $\widehat{C_{p}}$.

\begin{figure}[h]
    \centering
    \includegraphics[width=7cm]{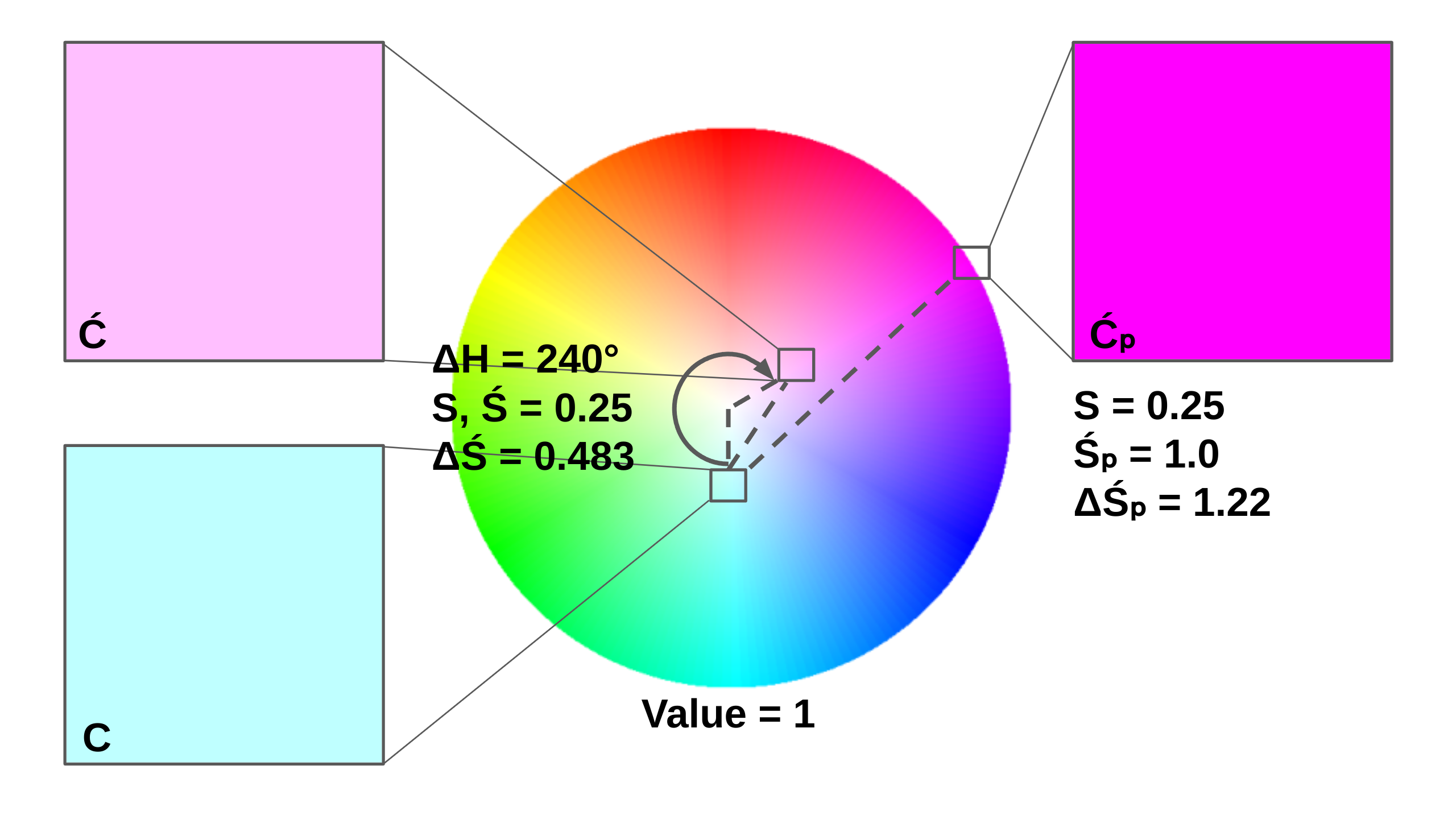}
    \caption{A color $C$, its ternary $\acute{C}$, and pure ternary $\acute{C_{p}}$ in HSV color space.}

    \label{fig:hsv_ternary}
\end{figure}

Additionally, we choose to introduce \textit{ternary} $\acute{C}$ and \textit{pure ternary} $\acute{C_{p}}$ colors of $C$, represented mathematically as
\begin{equation}
    \acute{C} = \{ (H - 120) \ \% \ 360, S, V\}
\end{equation}
\begin{equation}
    \acute{C_{p}} = \{ (H - 120) \ \% \ 360, 1, 1\}
\end{equation}

A ternary color can be plainly described as a color with a hue offset of $240^{\circ}$ clockwise from another color. A pure ternary color has the same hue offset, but its saturation and value must both be one. Figure \ref{fig:hsv_ternary} visualizes this relationship.

Relative to $C$, the Euclidean distance to $\acute{C}$ will be
\begin{equation}
    \Delta \acute{S} = \sqrt{(\cos(H_{t})\acute{S})^{2} + (S + \sin(H_{t})\acute{S})^{2}}
    \label{eqn:acute_s_1}
\end{equation}
with $H_{t} = 360 - \Delta H$. Notice, $\Delta H = 240 ^{\circ}$ for all values $H$ and $\acute{H}$ so $H_{t}=120^{\circ}$. Simplifying (\ref{eqn:acute_s_1}) yields
\begin{equation}
    \Delta \acute{S} = \sqrt{(-\acute{S}/2)^{2} + (S + \acute{S}\sqrt{3}/2 )^{2}}
\end{equation}
Similarly, the Euclidean distance from $C$ to $\acute{C_{p}}$ can be expressed as
\begin{equation}
    \Delta \acute{S_{p}} = \sqrt{(-1/2)^{2} + (S + \sqrt{3}/2)^{2}}
\end{equation}
since $\acute{S}=1$ for \textit{pure} colors.

To maximize marker visibility in varying color backgrounds, we need not only a high contrast in hue, but also a high contrast in saturation and value. In other words, for any background color in the 3D HSV color space, the marker color should be the one which is the furthest in Euclidean distance from the background color. However, active markers on the DS should emit with the highest intensity possible to extend viewing distance, so HSV marker colors must lie in the 2D HSV color space with value 1. \bc \textit{Therefore, the theoretical ideal marker color choice in the HSV color space is the pure complement of the background color.} \ec

Another consideration must be taken into account -- when the markers are emitting light underwater, the entire range of saturation and value is not reachable due to the attenuation of light and superposition described in Eq. \ref{eqn:superposition_light}. Therefore, the actual observed color separation between a color and its pure complement may not always be at least one. \bc Moreover, colors not in the background may be present on the DS itself, such as orange buoys, which can lead to false positives. \ec
As such, two different marker color mappings are evaluated: a pure complementary and pure ternary mapping.

\subsection{Vision Pipeline}

\begin{figure*}[ht]
    \centering
    \includegraphics[width=\textwidth]{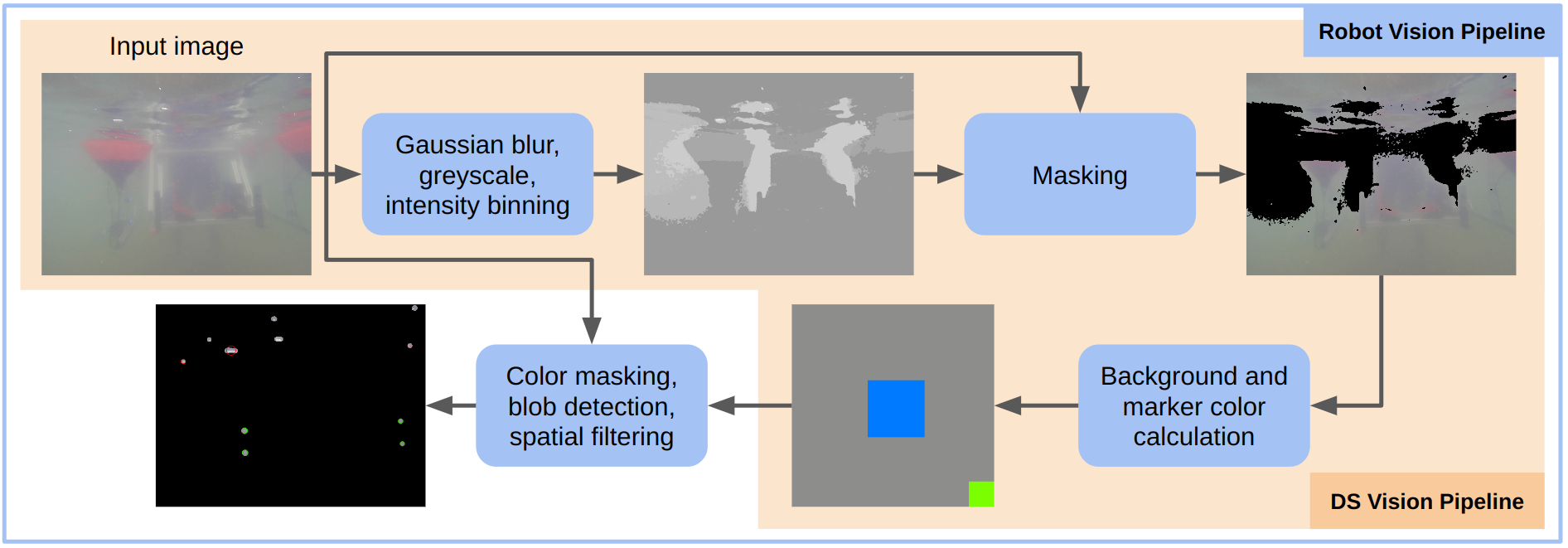}
    \caption{Vision pipeline for adaptive marker color configuration on the DS and robot. \bcc Algorithms are run on-board the machines in real-time. The robot pipeline shares common components with the DS pipeline until color landmark detection is necessary. \ecc}
    \label{fig:vision_pipelines}
\end{figure*}

In this section, we describe the algorithm used to determine DS marker color and detect markers. A flow chart of the vision pipeline for the AUV and DS can be seen in Fig. \ref{fig:vision_pipelines}. The pipeline is identical for both AUV and DS, except for landmark detection, which only runs on the AUV. Each respective pipeline runs in real time on a Raspberry Pi 4 and an NVIDIA Jetson TX2.

The AUV and DS initially read in an image from their camera to process. In order to determine the marker color that has the highest visibility in the current environment, the background color should be extracted from the image. The camera on the DS is collinear with the AUV camera, allowing both cameras to capture images of the same background. However, the AUV's view is obstructed by the DS, so any foreground objects in the camera's view should be removed. 

The image is first smoothed with a Gaussian blurring process to reduce noise and \bc smooth filter \ec responses further in the pipeline. The image is then converted from RGB to single channel greyscale and the pixels are binned into five equally spaced intervals from 0 to 255. The output of this step is a greyscale image with up to five different shades of intensities. The interval with the most number of pixels is assumed to contain only the background. A mask is created from this bin and applied to the original image. The resulting output image now contains the original color of the background with the foreground objects removed. The images containing the DS have a fairly uniform background color which increases the efficacy of this background extraction method.

The average background color $C_{b}$ is calculated in the HSV colorspace using the masked image. Two different marker color mapping functions are tested: pure complementary $\widehat{C_{pb}}$, and pure ternary $\acute{C_{pb}}$ mapping.

Once the marker color is determined for the frame, it is added to a two second rolling average. This reduces noise caused by the camera bobbing on the surface of the water.

The pipeline now diverges between the DS and AUV. The DS changes the color of the markers to the latest rolling average marker color. The color is converted back to RGB to interface with the common anode RGB LEDs and a fixed color correction offset is added to each RGB channel. 

According to the underwater imaging model in Eq. \ref{eqn:superposition_light}, the marker color \bcc sensed \ecc by the AUV will not be the color emitted from the DS marker, but a combination of the scattered components and the marker emitted light. Let $C_{AUVmkr}$ be the color of the DS marker \bc sensed \ec by the AUV's camera, $C_{bkg}$ the average background color from the background masked imaged, and $C_{mkr}$ the actual marker color, with all colors in the RGB color space. An approximation of the \bc marker \ec color light received can be expressed in the RGB color space as

\begin{equation}
    \begin{split}
        & C_{AUVmkr} = \{R, G, B\},\ given \\
        & R = max(R_{bkg}, (R_{bkg} + R_{mkr})/2)  \\
        & G = max(G_{bkg}, (G_{bkg} + G_{mkr})/2)  \\
        & B = max(B_{bkg}, (B_{bkg} + B_{mkr})/2)  \\
    \end{split}
\end{equation}

Since we can consider the RGB color space additive, each channel \bc must \ec have at least as much intensity as the background color, since the emitting marker cannot remove intensity. We also consider the forward scattering of the emitter negligible. For example, an emitter that is turned off (actual emitted RGB color $\{0, 0, 0\}$) in an underwater image will similar to the background color of the water. 

If the calculated marker intensity for a given channel is larger than the background color, an approximation for the color is calculated as the average between the background channel and calculated marker channel. In actuality, this value will increase as the AUV moves closer to the marker, and decrease as the robot moves further away, until it reaches a minimum of the background channel. Though inexact, this approximation is suitable for our needs.

Once $C_{AUVmkr}$ is calculated, this color is converted back into the HSV color space to specify the bounds for color masking. Color masking allows pixels in a range of hue, saturation, and value to pass through the pipeline, while blocking all other pixels that do not fall into these ranges. For the color $C_{AUVmkr}$, the mask range $C_{mask}$ can be represented in the HSV color space by

\begin{equation}
    \begin{split}
        & C_{mask} = \\
        & \{ [(H_{AUVmkr} \pm 60) \% 360], S_{mask}, V_{mask} \}\ with \\
        & \bc S_{mask} = [S_{AUVmkr} \pm 0.1765]\ s.t.\ 0 \le S_{mask} \le 1 \ec \\
        &\bc V_{mask} = [V_{AUVmkr} \pm 0.1765]\ s.t.\ 0 \le V_{mask} \le 1 \ec \\
    \end{split}
\end{equation}

Even though these are quite wide intervals, the filtered images contain relatively little noise. Since the color thresholding is allowing pure complementary and ternary colors of the background through, the portions of the image that remain are mainly markers, reflections of markers, and portions of the DS. Interval ranges were obtained experimentally.

\bcc The original image is masked with the inverse of the background mask to extract only foreground objects, then is color masked with the intervals described above. Next\ecc, a Laplacian of Gaussian (LoG) blob detection process is applied to extract round shapes from the image. Blobs are filtered by minimum area and are discarded if they are below the threshold. Finally, a spatial filter is applied to the blobs by using prior knowledge of the DS landmark locations. Any three blobs that form a right triangle with leg ratios similar to that of the rectangle side ratio formed by the DS markers are passed through. Any blobs that do not fit this criteria are discarded. Once four blobs that fit this criteria are found, point correspondence between the DS markers and the blobs is made by matching the top left, top right, bottom left, and bottom right most markers with the respective DS marker. The pose of the DS can now be calculated by solving the PnP problem \bc with a modern solver \cite{lepetit}. \ec

\section{Experiments}

Tests with the LoCO AUV and prototype DS were conducted off the shore of a lake and in a pool. The ternary color mapping function was tested in both environments, while the complementary mapping was only tested in the pool. Lake experiments were conducted by holding the AUV and DS predetermined distances apart from each other without obscuring visibility and allowing each system to float naturally in the water. Water depth ranged from approximately \bc 0.75 to 1.25 meters. \ec The DS and AUV were positioned 1, 3, and 5 meters apart and ran their marker color pipelines for approximately two minutes per distance interval. Data was recorded at each distance four times, with the robot facing north, east, south, and west to account for lighting variances. Water visibility was poor, which resulted in no DS features visible past approximately two meters. The pool experiment was conducted in a similar manner, but two tests, one for each color mapping function, were run per distance. The 1m complementary test distance drifted between 1 and 3m as the swimmer floated in the pool. \bc The auto exposure on the DS camera over corrected the image in the pool, but even so, the filtering range is robust enough to still detect the landmarks. \ec Lighting in the pool was uniform enough to waive the need to rotate the experimental setup. 

\begin{figure}[H]
    \centering
    \includegraphics[width=\textwidth]{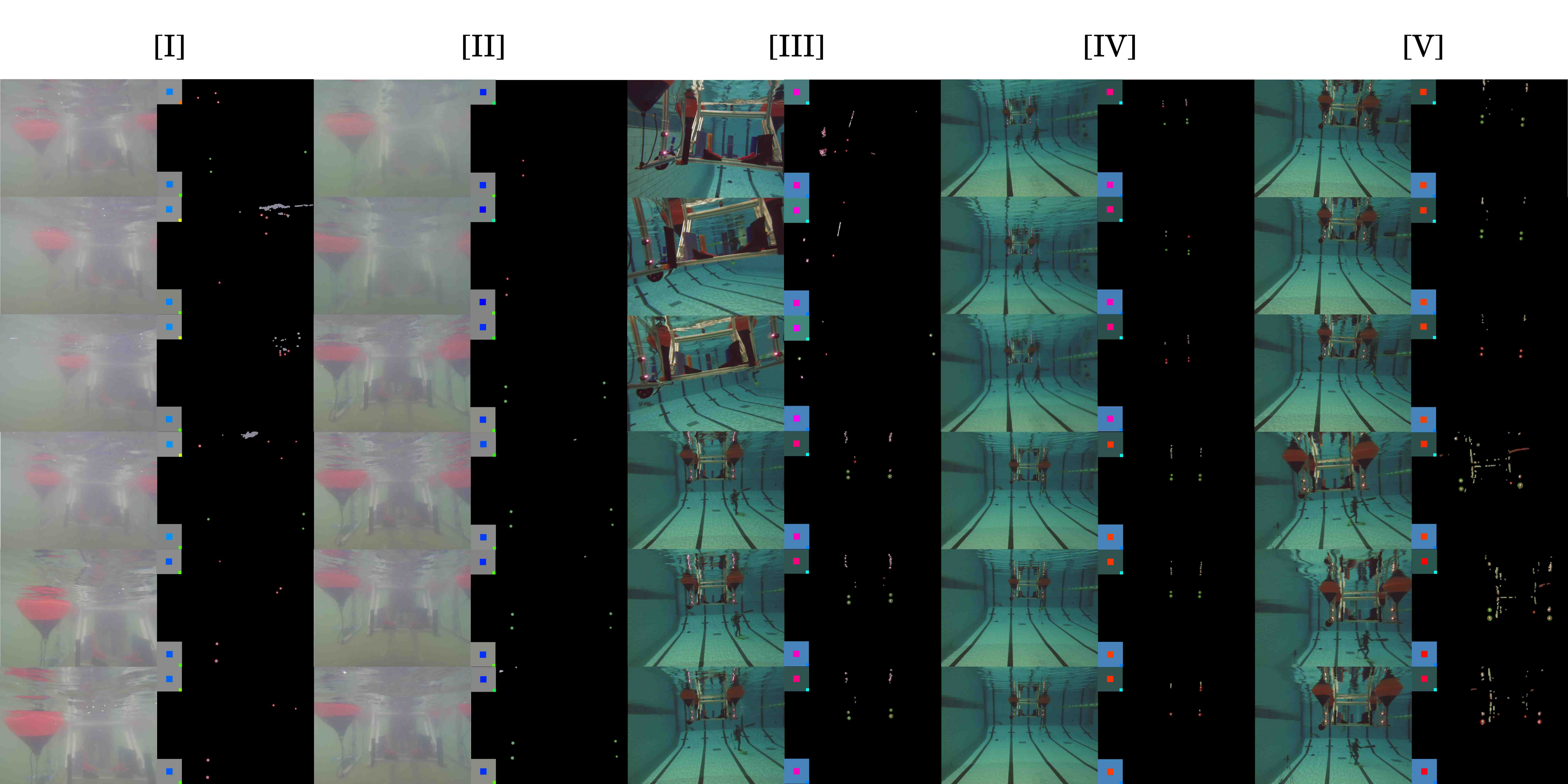}
    \caption{\bcc A sample of landmark detections from lake and pool environments. Lake data is from a distance of 1m. By column -- \textbf{[I]:} Top 4 Lake N, Bottom 2 Lake E \textbf{[II]:} Top 2 Lake W, Bottom 4 Lake S \textbf{[III]:} Top 3 Pool 1m Ternary, Bottom 3 Pool 3m Ternary \textbf{[IV]:} Top 3 Pool 5m Ternary, Bottom 3 Pool 5m Complement \textbf{[V]:} Top 3 Pool 3m Complement, Bottom 3 Pool 1m* Complement. See Fig. \ref{fig:single_eval} for details. Zoom in on the PDF for the better viewing. \\\hspace{\textwidth}*Intended distance is 1m, actual is between 1 and 3m.
    \ecc }
    \label{fig:marker_eval}
\end{figure}

\begin{figure}[ht]
    \centering
    \includegraphics[width=10cm]{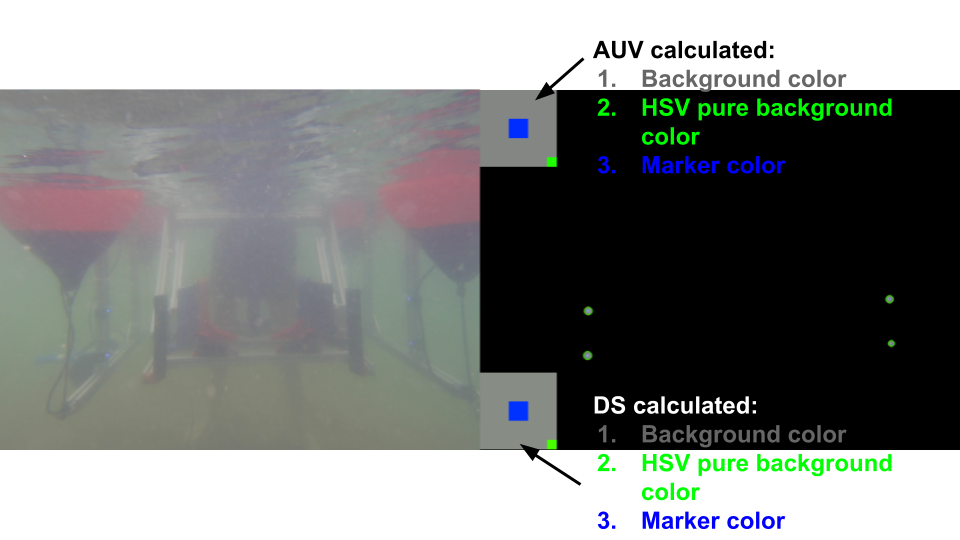}
    \caption{A side-by-side view of the raw camera image from the AUV on the left, and the masked and filtered marker image on the right. Green circles are detections that pass blob and spatial filtering, red circles pass only blob filtering. The top color swatch represents the background color, pure HSV background color, and marker color calculated by the AUV. The bottom swatch represents the same values calculated independently by the DS.}
    \label{fig:single_eval}
\end{figure}
\section{Results}

A \bcc qualitative \ecc summary of results can be seen in Fig. \ref{fig:marker_eval}. Each entry in this table consists of the raw AUV camera image, the calculated marker and background colors by both AUV and DS, and the final masked and filtered landmark image (refer to Fig. \ref{fig:single_eval} for details). Landmark detection is achievable even in low visibility lake conditions, as can be seen in columns I and II in Fig. \ref{fig:marker_eval}. Both DS and AUV independently agree on nearly the same marker color, which results in relatively little signal noise in the color masked image. Even with noise, the blob detection and spatial filtering methods correctly identify the true visible landmarks. The complementary filtering in columns IV and V show more false positives than ternary filtering. These false positives are from components on the DS itself, such as the orange buoys, metal struts, and 3D printed components.

\begin{figure}[H]
    \centering
    \includegraphics[width=\textwidth]{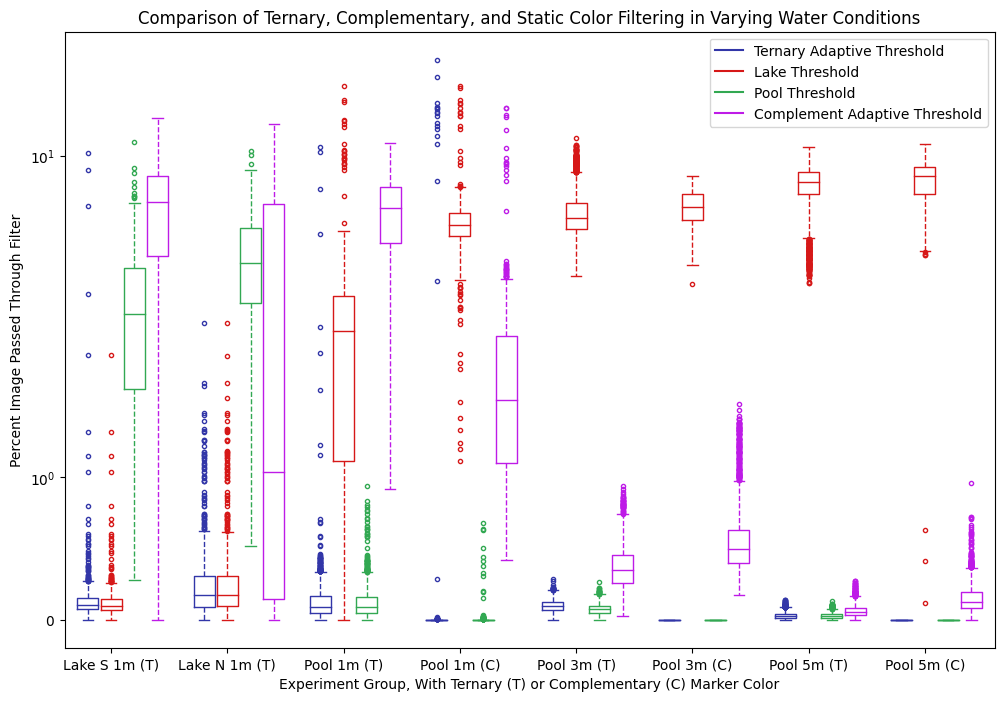}
    \caption{A box plot graph comparing static color thresholding with two cutoffs, one optimized for pool images and the other for lake, and adaptive ternary color thresholding. n $>$ 500 images per lake experiment and n $>$ 1500 images per pool experiment. The Y axis is log scale of the total percentage of the image passed through the filter, while the x axis is grouped into eight experimental runs.}
    \label{fig:comparison_graph}
\end{figure}

A quantitative comparison of adaptive and static color thresholding functions can be seen in Fig. \ref{fig:comparison_graph}. \bcc A key \ecc contribution of this work is to minimize the amount of noise allowed through a color filter in order to improve landmark detection and pose estimation. Therefore, we decide to perform our evaluation by comparing the total percentage of pixels passed through adaptive and static color filtering methods. Although it is not possible to tell which pixels are noise and which are true landmarks, minimizing the number of pixels that could possibly be landmarks is desirable to reduce the computational complexity of additional filtering, such as spatial filtering. The best performing filter will be the most selective and permit the fewest pixels through. 

\bcc As shown in Fig. \ref{fig:comparison_graph}, the ternary filter performs the best over all experimental setups. Image noise is over 10 times lower compared to static filters not tuned to the experimental environment. Almost zero noise is allowed through the ternary filter in experiments with complementary color DS markers, suggesting that pixels filtered through in ternary experiments are likely the DS markers. This is confirmed in the qualitative images. Static color filters perform well in their tuned environment, but are orders of magnitudes worse outside of the specific environment. As suggested by the qualitative results, the complementary filter picks up DS features other than markers, which allows large amounts of noise through even when DS markers display ternary colors. Overall, the ternary filter maintains highly selective filtering across all environments while successfully detecting DS landmarks.

\section{Conclusion}

This work shows that adaptive color thresholding with ternary colors is a robust method to detect a floating DS in visually dynamic aquatic environments. Water color can rapidly change due to a variety of environmental factors, causing previously contrasting landmarks to blend in with the background. Our proposed method of dynamically changing landmark color based on background water color is shown to perform over 10 times better than a static color filter landmark detector. Two adaptive color mapping functions were tested on the LoCO AUV and a prototype DS in two visually contrasting environments. Our pool and lake experiments show, qualitatively and quantitatively, that ternary color mapping provides the best noise rejection and detection of landmarks in visually varying marine environments.

\ecc

\end{document}